\documentclass{article}

\usepackage{siunitx}
\usepackage{booktabs}
\usepackage{multirow}
\RequirePackage[table]{xcolor}
\RequirePackage{ifpdf}
\RequirePackage{ifxetex}
\RequirePackage{ifluatex}
\RequirePackage{mathtools}
\RequirePackage{bm}
\RequirePackage{graphicx}
\RequirePackage{tabularx}
\RequirePackage{textcase}
\RequirePackage{dashrule}
\RequirePackage{ragged2e}
\RequirePackage{authblk}
\RequirePackage[hyphens]{url}
\RequirePackage{soul}
\RequirePackage{xpatch}
\usepackage[numbers]{natbib}
\title{Performance metrics for intervention-triggering prediction models do not reflect an expected reduction in outcomes from using the model}

\newcommand{\cf}[2]{#1^{(#2)}}
\newcommand{\Y}[1]{\cf{Y}{#1}}
\newcommand{\vY}[1]{\cf{\vec Y}{#1}}
\newcommand{\Yt}[1]{\cf{Y_t}{#1}}
\newcommand{\X}[1]{\cf{X}{#1}}
\newcommand{\Xt}[1]{\cf{X_t}{#1}}
\newcommand{\vX}[1]{\cf{\vec X}{#1}}
\begin{document}

\author{Alejandro Schuler \and Aashish Bhardwaj \and Vincent Liu}

\maketitle

\begin{abstract}
Clinical researchers often select among and evaluate risk prediction models using standard machine learning metrics based on confusion matrices. However, if these models are used to allocate interventions to patients, standard metrics calculated from retrospective data are only related to model utility (in terms of reductions in outcomes) under certain assumptions. When predictions are delivered repeatedly throughout time (e.g. in a patient encounter), the relationship between standard metrics and utility is further complicated. Several kinds of evaluations have been used in the literature, but it has not been clear what the target of estimation is in each evaluation. We synthesize these approaches, determine what is being estimated in each of them, and discuss under what assumptions those estimates are valid. We demonstrate our insights using simulated data as well as real data used in the design of an early warning system. Our theoretical and empirical results show that evaluations without interventional data either do not estimate meaningful quantities, require strong assumptions, or are limited to estimating best-case scenario bounds.

% % Please include a maximum of seven keywords
% \keywords{early warning system, predictive model, causal inference, model evaluation}
\end{abstract}

\section{Introduction}

Predictive models are now being deployed across healthcare settings to assist in diagnosis, prognosis, and risk stratification. In particular, these models are being used to trigger interventions for high-risk patients in hospital settings. While some risk models deliver a single prediction at a static timepoint during an inpatient encounter (e.g., at hospital admission), other models deliver risk scores or alerts repeatedly through time. For example, the Advanced Alert Monitor described by \citet{kipnis2016}, provides hourly scores that estimate the risk that an inpatient will deteriorate within the next 12 hours. If the risk is above a 5\% threshold, the system fires an alert that, after clinical review, can trigger a rescue intervention. Similar risk scores have been developed that are designed to initiate interventions to prevent sepsis, kidney injury, and other adverse outcomes \cite{prytherch2010, churpek2014, cretikos2007, smith2008, desautels2016, cheng2017, meiring2018, reyna2019, Henry:2015kn, tomasev2019}. Most of these systems have evolved from traditional acuity or risk scores that were manually calculated and used intermittently, but are now capable of producing real-time scores by leveraging electronic health record data. \cite{smith2008, legall2005, wilson1990}.

The ultimate goal of these risk model-triggered alert systems is to reduce the incidence of an adverse outcome, while also minimizing the number of necessary interventions \cite{pierce1884,vickers2006}. Thus, for two models that have an identical potential for mitigating an adverse outcome, the model that results in fewer interventions would be more desirable. Ideally, each model's benefit would be assessed through randomized clinical trials. Clinical trials for these systems are particularly appealing because the alert system could be active in the intervention arm, while it could be suppressed in the control arm. The outcome rates among the two groups could then be compared to assess the change in outcome rates attributable to the model-triggered intervention. To account for the clinical workload of the system, the total number of alerts (per patient, per day, etc.) would be also be tallied. The final utility of the system could be determined using a tradeoff between the number of prevented outcomes at the cost of the number of incurred interventions, depending on how these are valued by stakeholders \cite{vickers2016, vickers2006, vickers2008, pierce1884, baker2009, wu2016, reyna2019}. Multiple models could also be compared in this way by randomizing which model triggers the intervention during a given encounter. %Evaluations of this kind have been performed in a limited fashion for some models, with mixed results \cite{}.

Unfortunately, large randomized trials may not be feasible for every proposed model-based alert system, either due to cost, time, or uncertain equipoise. Nonrandomized prospective evaluations are possible, but these also require model deployment. Because of these limitations, researchers frequently rely on retrospective data for model evaluation or alert threshold selection. However, many prediction model developers also fail to consider the specific intervention that their models will actually trigger. Since the intervention is not known a-priori, we can only assume that it is not present in our retrospective data. While some models may trigger an existing intervention, if we cannot specify what it is, we must assume it is a new, as-of-yet unimplemented intervention that is not present in our data. The inherent assumptions and implications of using non-interventional retrospective data for model evaluation have been poorly addressed, particularly in the repeated-prediction setting. A number of different evaluation strategies have been used without using a framework for critical assessment or comparison. Our primary goal is to uncover what, if anything, these evaluation strategies are actually estimating and to offer guidance on how to best evaluate repeated-predictions models. 

\subsection{Outline}

For models that deliver a single prediction per independent patient (or encounter, etc.), standard performance metrics derived from confusion matrices (e.g. sensitivity) calculated using retrospective data are often used to describe clinical utility in terms of preventable outcomes and additional workloads. However, it is not commonly appreciated that these performance metrics are merely proxies for clinical utility under certain assumptions, and that maximizing them should not be the end goal in and of itself \cite{Ascarza:2018ie, kleinberg2015prediction, athey2017beyond}. We will therefore first review the basis of these evaluation metrics for single-prediction models and demonstrate how they are used as proxies for clinical utility.

After doing this, we will move on to evaluation schemes for models that deliver repeated predictions (e.g. multiple scores throughout a single hospitalization). The lack of an a-priori theoretical justification for evaluation has led to several methods described in the literature. In fact, for most of these evaluations, what exactly is being estimated is poorly described. Using a unified theoretical framework, we will review these differing approaches and demonstrate why it is not trivial to extend approaches based on confusion matrices to repeated-predictions models.

\section{Evaluating Single-Prediction Models}

For prediction models that deliver a single prediction per encounter or patient, standard evaluation metrics based on the confusion matrix of the predictions and outcomes in retrospective data (a.k.a. the 2x2 table or contingency table) are reasonably useful proxies for the anticipated reduction in outcomes and workload. We first describe this heuristically and then bolster these concepts with mathematical rigor.

Consider, for instance, a prediction model that runs at the time a baby is born and which triggers an alert if the baby's risk of developing sepsis in the next week exceeds 5\%. We are completely agnostic to how this model was trained in the first place. However, using unseen retrospective data, we can tally the number of cases where the alert would have fired and the baby did go on to develop sepsis- these would be our "true positives". Similarly, we could calculate the false positives, false negatives, and true negatives. These counts constitute the four cells of the familiar confusion matrix, which can the be used to calculate metrics like sensitivity, specificity, and precision. 

For single-prediction models, the counts in the confusion matrix are, under relatively simple assumptions, related to we are interested in: how many outcomes could we potentially prevent, and for how many alerts. 

The number of true positives represents the \textit{maximum} number of outcomes that we could ever hope to prevent with the interventions that follow the alert. In almost every scenario, because interventions are not perfectly effective, fewer actual outcomes will be prevented. Some outcomes will not be preventable, even if they were foretold (i.e. the true positives overestimate the number of prevented outcomes). Conversely, outcomes that did occur and were not foretold would still have occurred without the alert system (i.e. no false negatives can be prevented by the alert system). Thus, the number of true positives is an upper bound on the number of outcomes the alert system could prevent. 

However, the upper bound is still useful for comparison under the assumption that there is no heterogeneity in the effect (i.e. the intervention equally reduces everyone's chance of an outcome). For instance, assume that an intervention triggered by the alert has a 50/50 chance of preventing an outcome that otherwise would have happened, regardless of the patient. Thus, if model A produced 100 true positives it would translate into 50 prevented outcomes, whereas for model B which produces 80 true positives, it would only translate into 40 prevented outcomes. Under these assumptions, model A would be deemed superior because it could prevent more outcomes. This relationship holds even if the percentage of outcomes prevented by the intervention (i.e. the treatment effect of the intervention) is unknown. If the treatment effect is constant across patients, the ratio of the true positives between two models would still correspond to the ratio of prevented outcomes. 

Note, however, that this may not be the case, so statistics calculated from these counts (e.g. sensitivity, specificity) may not actually track with the clinical utility of a model, even for single-prediction models. Consider two models that both have 100 true positives and 100 false negatives. Imagine that there are two kinds of high-risk patients: patients who are high-risk, but won't respond to treatment (perhaps they are "too far gone") and patients who are high-risk but will respond. If one model correctly classifies the high-risk, low-response patients, but not the high-risk, high-response patients, its utility will be far worse than a model that correctly classifies the high-risk, high-response patients, but not the high-risk, low-response patients. This happens despite the fact that both of these models have the same number of true positives and false negatives (thus the same sensitivity). In other words, if patients who are most at risk are not necessarily those who may benefit the most from intervention, standard metrics will not track with model utility \cite{Ascarza:2018ie}.

The total number of alerts incurred by a single-prediction model-based alert system is also easily calculated from retrospective data: it is the total number of predicted positives- the false positives plus the the true positives. This number is exact in the sense that if an alert would have fired in retrospective data, we know it would have fired in the same scenario prospectively. 

\subsection{Potential Outcomes}

These ideas can be pinned down with some mathematical notation in the potential outcomes framework \cite{Rubin2005}. Let $W$ be the alert status (and thus the intervention), and $X$ a vector of predictors or covariates that are used in a model $f$ to trigger the alert (i.e. $W = f(X)$). Let $Y^{(0)}$ be the outcome that we would have observed had the intervention not occurred and let $Y^{(1)}$ be the outcome had the intervention been triggered. When $f$ is used to control the alert, the observed outcome is $Y = Y^{(0)}(1-f(X)) + Y^{(1)} f(X)$. In other words, when the intervention happens, we observe $Y^{(1)}$, and when it doesn't, we observe $Y^{(0)}$. The number of outcomes among $n$ patients when using the model $f$ to trigger alerts is $nE[Y] = nE[Y^{(1)} f(X) + Y^{(0)} (1-f(X))]$ where the expectation is over the entire eligible population. The number of alerts among $n$ patients is $nE[W] = E[f(X)]$.

\subsubsection{Estimating Prevented Outcomes}

The expected number of outcomes (per patient) when using a model $f$ to trigger alerts is

$$
\begin{array}{rcl}
E_f[Y]/n
&=& E[\Y{1} f(X) + \Y{0} (1-f(X))] \\
&=& E[(\Y{1} - \Y{0})f(X) + \Y{0}] \\
&=& E[E[(\Y{1} - \Y{0})f(X)|X=x]] + E[\Y{0}] \\
&=& E[(\cf{\mu}{1}(X) -  \cf{\mu}{0}(X))f(X)] + E[\Y{0}]
\end{array}
$$

where $\cf{\mu}{1}(x)$ and $\cf{\mu}{0}(x)$ are the conditional means of the potential outcomes under the intervention and no intervention conditions, respectively. If the ratio between these two is a constant risk ratio $\rho = \cf{\mu}{1}(x)/\cf{\mu}{0}(x)$ for all $x$, then we have

$$
\begin{array}{rcl}
E_f[Y]/n 
&=& E[(\cf{\mu}{1}(X) -  \cf{\mu}{0}(X))f(X)] + E[\Y{0}] \\
&=& (\rho- 1) E[\cf{\mu}{0}(X) f(X)] + E[\Y{0}] 
\end{array}
$$

These expectations are easily estimated using retrospective data. The key observation is that we can treat the retrospective data (including predicted alerts $W$ as samples from the joint distribution of $(\Y{0}, W, X)$. We only ever observe $\Y{0}$, regardless of $W$, because in the retrospective data no alert ever fired and thus no intervention was actually delivered.

Because of this, $E[\Y{0}]$ is the rate of outcomes observed in the retrospective data $E[Y]$ and is thus estimated by $n_{Y=1}/n$, the empirical rate of outcomes in the retrospective data. $E[\cf{\mu}{0}(X) f(X)]$ is the long-run average of how many times the outcome co-occurred with an alert in the retrospective data, so it is estimated by $n_{Y=1, W=1}/n$, the empirical rate of true positives in the retrospective data.

Thus an unbiased estimate of the number of outcomes expected when using the model $f$ is 

$$
n_{Y=1} - (1-\rho)n_{Y=1,f(X)=1}
$$

$n_{Y=1}$ is the total number of outcomes observed in the retrospective data, meaning that using the model $f$ would prevent $(1-\rho)n_{Y=1,f(X)=1}$ outcomes among $n$ patients prospectively. The risk ratio $\rho$ is not known in general, but the best-case scenario is $\rho=0$, meaning that an estimated $n_{Y=1,f(X)=1}$ outcomes would be prevented among $n$ prospective patients. This is an upper bound that holds no matter what $\rho$ actually is, and even if there is heterogeneity of effect. It is thus a useful, relatively assumption-free quantity.

However, if we are willing to assume a constant population-wide risk ratio, the number of true positives also facilitates the comparison of two models on the basis of prevented outcomes (not an upper bound), even if we do not actually know what that risk ratio is. Proceeding from above, the estimated difference in outcomes between two models $f$ and $g$ both triggering the same intervention with risk ratio $\rho$ is

$$
(\rho-1)(n_{y=1,f(x)=1} - n_{y=1,g(x)=1})
$$

The (possibly unknown) risk ratio $\rho$ is fixed, so the difference in estimated prospective outcomes depends only on the difference in the number of observed true positives in the retrospective data. That means that the number of outcomes prevented by a model will track with its estimated sensitivity as long as the risk ratio of the intervention is assumed to be constant across the population.

\subsubsection{Estimating Alert Workload}

The number of prospectively incurred alerts is estimated using the empirical number of alerts that would have fired in the retrospective data because the distribution of covariates in the retrospective and prospective data are presumed to be the same. So $nE[W] = nE[f(X)]$, which is estimated by $n_{W=1} = \sum w$. 

\section{Evaluating Repeated Prediction Models}

Instead of predicting at a fixed point in time, many authors are now considering models that deliver predictions repeatedly over a patient encounter. For instance, a model might predict a patient's 12-hour risk of kidney injury at every hour during their hospital stay. Again, we make no assumption about how these models may be trained, but we assume that we have access to a previously unseen retrospective sample of data (i.e. a test set).

The evaluation of models that make repeated predictions is more complicated than for models that make single predictions, although to our knowledge this has not been explicitly pointed out in the literature. The complications in this setting are that the relationship between intervention and outcome is mediated through time and that interventions can affect the trajectory of a patient, making the portions of the retrospective data after an initial alert unrepresentative of what would be observed prospectively.

Several approaches have been used to evaluate models of this kind. 

\citet{cheng2017}, \citet{avati2018}, and \citet{churpek2014} do not specify a-priori when an alert will fire. Approaches of this kind do not accurately estimate any relevant quantity and would be impossible to apply prospectively because it is unclear when the model would be used to trigger an alert and when it would not. For instance, for patients who experienced an outcome \citet{cheng2017} use their model to predict a certain number of days before that outcome. Prospectively, however, it would be impossible to know which patients would go on have an outcome and, if so, exactly when. It would thus be impossible to use the model in this way. We will not further discuss these approaches and do not recommend their use.

\citet{koyner2016} and \citet{prytherch2010} use an \textit{aggregated time} evaluation, in which each timepoint of retrospective data is labeled as a true positive, false negative, etc. depending on whether an alert fired in that timepoint and whether or not an outcome occurred in the proceeding lookahead period. Although it appears to do so, this approach may not correctly estimate quantities like the positive predictive value of an alert (i.e., given that there was an alert for a random patient at a random timepoint, how likely is it that an outcome occurs in the next lookahead period?). The reason for this failure is that the joint distribution of alerts and outcomes in the retrospective data is generally not the same as that of the prospective data, since, once an alert occurs prospectively, all future quantities are presumably affected.

\citet{shickel2019}, \citet{meiring2018}, and others \cite{desautels2016, wilson1990} perform separate evaluations of their model at each timepoint, an approach that we will call the \textit{fixed time} evaluation. This approach is essentially the single-prediction evaluation strategy, which assumes that the repeated-prediction model will actually only be used at a single timepoint. However, if the model is used at multiple timepoints in practice, the performance characteristics at timepoints past the first will not be accurately reflected by the estimates.

\citet{Henry:2015kn} count the number of outcomes in any period after an alert as true positives and do not count any alerts after the first towards the alert total. We call this the \textit{first alert} evaluation method. This estimates an upper bound on the rate of preventable outcomes and accurately estimates the rate of prospectively incurred alerts, assuming all alerts for a patient will be ignored after the first. 

\subsection{Potential Outcomes}

What follows is all borrowed from the causal inference literature on time-varying treatments \cite{whatif}. Let $W_t$ represent the alert at timepoint $t$. For simplicity, we will assume a fixed, maximum timepoint denoted $T$ for what follows. Denote a trajectory of measurements up to but not including time $t$ as $\vec A_t = [A_0, A_1, \dots A_{t-1}]$. At each point $t$, there are now $2^t$ possible potential outcomes. For instance, at time $t=2$, there were either no preceding alerts ($\vec W_2=[0,0]$), or there was an alert at the first timepoint ($\vec W_2=[1,0]$), or the second ($\vec W_2=[0,1]$), or both ($\vec W_2=[1,1]$). We denote the potential outcome at time $t$ corresponding to treatment $\vec W_t$ using the notation $\Yt{\vec W_t}$. The values of the covariates also now have counterfactual distributions, which we denote $\Xt{\vec W_t}$.

We must assume that the outcomes $Y_t$ and covariates $X_t$ that we observe at each time in the retrospective data are the potential outcomes $\Yt{\vec 0_t}$ and covariates $\Xt{\vec 0_t}$ corresponding to the alert trajectory $W_t = [0, 0, \dots 0]$, since no alerts were actually fired in this data. 

In the prospective setting, we would apply the alert system triggered by the model $f$. Because the alerts are triggered by a predictive model $f$, we have a dynamic treatment strategy that complicates the construction of what the observed data would be in terms of the counterfactuals. For instance, consider the covariates we observe at time $t=1$ given that $f$ is used to trigger the alerts. In our notation, that quantity would be $\X{[f(X_0)]}_1$. Note that this random variable is not distributed as the potential covariate $\X{[0]}_1$ nor as the potential covariate $\X{[1]}_1$. It is a mixture of the two:

$$
\X{[f(X_0)]}_1 
= 
\X{[0]}_1  1_0(f(X_0)) + 
\X{[1]}_1  1_1(f(X_0))
$$

% simulation would clarify this point 

Now consider the covariates at time $t=2$, which would be $\X{[f(X_0), f(X_{1,[f(X_0)]})]}_2$. This is an even more complex mixture. The notation is extremely cumbersome at this point, so we will denote these quantities at time $t$ using the simplified notation $\Xt{f}$ (and $\Yt{f}$ for the outcomes). Do remember, however, that these variables are complex mixtures that depend on the contemporaneous potential variables and the preceding potential variables.

The fundamental problem we face when using retrospective data to evaluate an alert system is that the data we have are samples from $({\vX{\vec 0}}, {\vY{\vec 0})}$, whereas we are interested in estimating quantities we would observe in the counterfactual universe $(\vX{f}, \vY{f})$ in which alerts are being triggered by a model $f$. 

This is at heart the same problem that exists in the single-prediction setting, where we observe $(X, \Y{0})$ but are interested in $(X, \Y{f(X)})$. However, in the single-prediction setting, postulating a simple relationship between $\Y{1}$ and $\Y{0}$ is enough to allow calculation of all relevant counterfactual quantities. In the repeated-prediction setting, future patient covariates and outcomes depend on the history of past interventions, meaning that a) there are many, many more possible counterfactual universes that need to be related to each other by some assumptions and b) the relationships between these counterfactuals are themselves much more complicated because of the dynamics through time.

The retrospective data, then, are largely inadequate to support a meaningful evaluation of the model. As we will see, there are many summary statistics that may be calculated using these data, but they may not have clear and useful interpretations in terms of an expected reduction in outcomes or an expected alert workload.

% The only dependence structure we assume is that concurrent measurements cannot affect each other and that future measurements cannot affect past measurements. However, past alerts can (and presumably should) impact future covariates (and thus future alerts). 

% In this context, the (per patient) alert rate of a model $f$ is

% $$
% \alpha(f) = 
% E\left[
% \sum_{t=0}^T f(X_{t,f}) 
% \right]
% $$

% and the outcome rate is

% $$
% \gamma(f) = 
% E\left[
% \sum_{t=0}^T Y_{t, f}
% \right]
% $$

\paragraph{Aggregated Time}

The aggregated time evaluation uses the retrospective data at all timepoints to populate a confusion matrix. The "estimate" is the value of the alert at any timepoint $W_t = f(X_t)$, and the "truth" is whether or not an outcome occurred in a lookahead period of a certain length: $\tilde Y_t = \sum_{\tau \in (t, t + \mathcal T]} Y_\tau$. The values for $W_t$ and $\tilde Y_t$ for each patient-timepoint are then used to tally up the number of true positives: $n_{W_t=1, \tilde Y_t=1}$, false positives: $n_{W_t=1, \tilde Y_t=0}$, etc. 

This evaluation is direct extension of how a standard machine learning model might be trained in order to predict an outcome within a lookahead period. In that sense, it is the evaluation that most researchers or data scientists trained in machine learning might immediately reach for. Unfortunately, however, the confusion matrix counts in this evaluation are not useful in estimating any meaningful quantity without heroic assumptions.

\begin{table}[ht]
    \centering
    \begin{tabular}{|c|c|c|c|c|}
        \hline
        Patient & Time & Alert? & Outcome in lookahead period? & label \\
        \hline
        1 & 1 & 0 & 0 & True Negative \\
        1 & 2 & 0 & 0 & True Negative \\
        1 & 3 & 1 & 0 & False Positive \\
        1 & 4 & 0 & 1 & False Negative \\
        1 & 5 & 1 & 1 & True Positive \\
        2 & 1 & 0 & 0 & True Negative \\
        2 & 2 & 0 & 0 & True Negative \\
        2 & 3 & 1 & 0 & False Positive \\
        \hline
    \end{tabular}
    \caption{Example data structure for the aggregated time evaluation}
\end{table}

For instance, it is tempting to treat the number $n_{W_t=1}$ as an estimate of $nE[\sum W^{(f)}_{t}]$, the total number of alerts that would fire among $n$ patients throughout their stays. However, this does not take into account the fact that the "observed alerts" in the retrospective data did not actually occur. If they had, all of the data subsequent to those alerts would have been affected and different from what was actually observed. $n_{W_t=1}$ is in fact an estimate of $nE[\sum W^{(\vec 0)}_{t}]$: the number of alerts that would fire among $n$ patients assuming that those alerts are "silent" and cannot affect future patient covariates that would determine future alerts. Therefore, unless the alerts do not affect future patient covariates (e.g. physiology), the number of alerts ``observed'' in the retrospective data will not be representative of the number that would be observed if the alert system were running. The number of "true positives" $n_{W_t=1, \tilde Y_t=1}$ is similarly unrepresentative of the number of outcomes that would have been potentially prevented in some lookahead period by an alert. 

Thus, without strong assumptions to relate the quantities $nE[\sum W^{(\vec 0)}_{t}]$ and $nE[\sum W^{(f)}_{t}]$ or $nE[\sum Y^{(\vec 0)}_{t}]$ and $nE[\sum Y^{(f)}_{t}]$, the numbers $n_{W_t=1}$ and $n_{W_t=1, \tilde Y_t=1}$ are not meaningful. Metrics such as sensitivity, precision, etc. derived from the aggregated time evaluation will therefore not relate to any useful notion of clinical utility.

\paragraph{Fixed Time}

The fixed time evaluation considers the retrospective data at a single timepoint $t=t^*$ to populate the counts in a confusion matrix. The "estimate" is the value of the alert at that time: $\tilde W = W_{t^*} = f(X_{t^*})$, while the "truth" is whether or not an outcome occurred at any point after that time: $\tilde Y = \sum_{t = t^*}^T Y_t$ (assuming, for simplicity, that the outcome can only occur once). The values for $\tilde W$ and $\tilde Y$ for each patient are then used to tally up the number of true positives: $n_{\tilde W=1, \tilde Y=1}$, false positives: $n_{\tilde W=1, \tilde Y=0}$, etc. 

\begin{table}[ht]
    \centering
    \begin{tabular}{|c|c|c|c|c|}
        \hline
        Patient & Time & Alert? & Outcome in future? & label \\
        \hline
        1 & 5 & 1 & 0 & False Positive \\
        2 & 5 & 1 & 1 & True Positive \\
        3 & 5 & 0 & 0 & True Negative \\
        4 & 5 & 0 & 1 & False Negative \\
        \hline
    \end{tabular}
    \caption{Example data structure for the fixed time evaluation}
\end{table}

In effect, the fixed time evaluation reduces a repeated-prediction model to a single-prediction model that is used at time $t^*$. Consequently, all of the arguments justifying the utility of the single prediction metrics are applicable to the fixed time evaluation as well. Denote $\cf{e_t}{w} = [0,0,\dots w, 0,0, \dots 0]$: a vector of length $t$ with the value $w$ at element $t^*$. In the fixed time evaluation, the only two possible treatment trajectories are $\cf{e_T}{0}$ and $\cf{e_T}{1}$, since the alert cannot fire either before or after time $t^*$. In the derivation of the outcome and alert rates, we can replace $Y$ by $\tilde Y$, $W$ by $\tilde W$ and the potential outcomes $\Y{w}$ with $\cf{\tilde Y}{w} = \sum_{t=t^*}^T Y_t^{\left( \cf{e_t}{w} \right)}$. The derivation then proceeds identically.

Therefore the total number of positives $n_{\tilde W=1}$ is an unbiased estimate for $nE[\tilde W] = nE[f(\X{\vec 0}_{t^*})]$. This is the number of patients for whom we expect alerts to fire out of $n$ patients \textit{at time $t^*$}, \textit{assuming the alert system was not turned on until that point in time}. Without different and likely much more heroic assumptions, $n_{W^*=1}$ does not estimate the alert count we would expect to observe prospectively at time $t^*$ if the alert system were turned on \textit{before} $t^*$, which would be $nE[f(\X{f}_{t^*})]$. It also does not represent the total alert count that would be incurred by the system if it were running at all times, which would be $nE[\sum_t f(\Xt{f})]$.

Similarly, $n_{\tilde W=1, \tilde Y=1}$ is the maximum possible number of outcomes that could be prevented by using the model $f$ to trigger an alert system \textit{at time $t^*$}, \textit{assuming the alert system was not turned on until that point in time}. This is not an estimate of the maximum reduction in outcomes we could expect if the alert system were running \textit{continuously}. And, as is the case in the single-prediction setting, it is not possible to estimate anything other than a bound on the reduction of outcomes without unsubstantiated assumptions about the relationship between $\cf{\tilde Y}{0}$ and $\cf{\tilde Y}{1}$. 

\subparagraph{Look-ahead Prediction Windows}

In the literature, the fixed-time evaluation is often used in conjunction with a lookahead window for the outcome. In other words the "truth" is only positive if the outcome occurred in some window of length $\mathcal T$ after $t^*$: $\tilde Y = \sum_{t \in (t^*, t^* + \mathcal T]} Y_t$. This outcome definition is then used to populate the counts in the confusion matrix. 

The problem with using look-ahead windows is that they can complicate the interpretation of the upper bound on the reduction in outcomes. If, ultimately, we are really only interested in reducing outcomes in a window of fixed length after $t^*$, then there is no issue: $n_{\tilde W=1, \tilde Y=1}$ is a fair estimate of the maximum number of outcomes we could hope to reduce in that window. If, on the other hand, we are ultimately interested in the \textit{total} number of outcomes prevented (say, throughout the hospitalization), the number of ``true positives'' from this evaluation is not useful. On the one hand, $n_{\tilde W=1, \tilde Y=1}$ is an overestimate of the true number of outcomes that would be prevented in the lookahead period (since it is an upper bound). On the other hand, it could also be an underestimate of the total number of outcomes prevented in the future, since some outcomes that are further out in time than $\mathcal T$ could also be prevented. Thus $n_{\tilde W=1, \tilde Y=1}$ is not conclusively either a lower or upper bound on the total number of outcomes prevented by alerting based on the model $f$ at time $t^*$, but not before. Without heroic assumptions about how the treatment effect of the intervention triggered by the alert varies in time, this is not an informative quantity. Sensitivity, precision, and other derived measures are thus likewise uninformative in the fixed time evaluation.
\\

In conclusion, the fixed time evaluation can accurately estimate the number of alerts that would be produced by a system that is turned on only for a given timepoint, but not before or after. It can also estimate an upper bound on the number of outcomes that such a system could conceivably prevent, but that bound is of questionable utility for comparison between models if the intervention is time-sensitive. In practice, the fixed-time evaluation is often repeated for each timepoint and the results compared in order to decide at what time the model should be run. The results are not useful if the model is meant to run continuously.

\paragraph{First Alert}

The first alert evaluation discards all alerts after the first and considers all outcomes after that alert to have been correctly predicted. Any alerts after an outcome are also discarded. This corresponds to the scenario where the alert system is running ``continuously'' (i.e. at each timepoint), but is turned off after any alert or outcome for a given patient. 

\begin{table}[ht]
    \centering
    \begin{tabular}{|c|c|c|c|c|}
        \hline
        Patient & Any alert? & Any outcome? & label \\
        \hline
        1 & 0 & 0 & True Negative \\
        2 & 1 & 0 & False Positive \\
        3 & 0 & 1 & False Negative \\
        4 & 1 & 1 & True Positive \\
        \hline
    \end{tabular}
    \caption{Example data structure for the first alert evaluation}
\end{table}

We can encode the condition that the alert system is turned off after a first alert or outcome by introducing a new alerting function 

$$
\tilde f(x_t, \vec w_t, \vec y_t)
= \begin{cases}
f(x_t) & \text{if}\  \vec w_t = \vec 0, \ \vec y_t=\vec 0 \\
0 & \text{else} 
\end{cases}
$$

But instead of complicating our notation, we will continue to use $f(x_t)$ when really we mean $\tilde f(x_t, \vec w_t, \vec y_t)$. 

For each observation, let a random variable $T_{W}$ denote the time of the first alert, if any, or $0$ if there was no alert. Then, in terms of the unobserved counterfactuals, the observed (retrospective) data are $\tilde W = \sum^{T_{W}} f(\Xt{\vec 0})$ and $\tilde Y = \sum^T \Yt{\vec 0}$ (again assuming for simplicity only a single outcome is possible per encounter). 

Our first result is that the number of observed first alerts $n_{\tilde W=1}$ is an unbiased estimate of the number of first alerts that would be observed in a prospective dataset with the same number of patients. The reason for this is that $\sum^{T_{W}} f(\Xt{\vec 0}) = \sum^{T_{W}} f(\Xt{f})$ because, by definition, $\Xt{f} = \Xt{\vec 0}$ when $t \le T_{W}$ (i.e. the conditioned random variable $\Xt{f}|t \le T_{W}$ is $\Xt{\vec 0}$). Assuming that alerts after the first are ignored or ``snoozed'', this is an unbiased estimate of the alert workload that would be incurred by using the model $f$ to trigger alerts.

The number of observed true positives $n_{\tilde W=1,\tilde Y=1}$ is an upper bound for the number of outcomes that could possibly be prevented by triggering alerts using the model $f$. Although the mathematics become somewhat convoluted, the intuition is simple: any outcomes that happen after a first alert would have gone off are outcomes that potentially could have been prevented. The proof is as follows.

The difference in outcomes between using the model $f$ to alert and issuing no alerts is

$$
\begin{array}{rcl}
nE\left[
    \sum_0^T \Yt{\vec 0} - \sum_0^T \Yt{f}
\right] 
&=& nE\left[
    \sum_0^{T_W} \Yt{\vec 0} - \sum_0^{T_W} \Yt{f} + 
    \sum_{t=T_W}^{T} \Yt{\vec 0} - \sum_{t=T_W}^T \Yt{f}
\right] \\
&=& nE\left[
    \sum_{t=T_W}^{T} \left(
        \Yt{\vec 0} - \Yt{f}
    \right)
\right] \\
\end{array}
$$

Abbreviate $Z = \sum_{0}^T f(\Xt{f})$. Note $Z \in {0,1}$ since either no alert fires or a single alert fires. Furthermore, note that the quantity $\sum_{t=T_W}^{T} \left( \Yt{\vec 0} - \Yt{f} \right) | Z=0$ is $0$ because $Z=0$ implies there were no alerts, meaning $\Yt{\vec 0} = \Yt{f}$ for all $t$. Then, continuing,

$$
\begin{array}{rcl}
&=& 
nE\left[
    \sum_{t=T_W}^{T} \left( \Yt{\vec 0} - \Yt{f} \right) | Z=0
    \right]P(Z=0) + 
nE\left[
    \sum_{t=T_W}^{T} \left( \Yt{\vec 0} - \Yt{f} \right) | Z=1
    \right]P(Z=1) \\
&=& nE\left[
    \sum_{t=T_W}^{T} \left( \Yt{\vec 0} - \Yt{f} \right) | Z=1
    \right]P(Z=1) \\
&=& nE\left[
    \sum_{t=T_W}^{T} \Yt{\vec 0} | Z=1
    \right]P(Z=1) - 
    nE\bigg[
    \sum_{t=T_W}^{T} \Yt{f} | Z=1
    \bigg]P(Z=1) \\\\
\end{array}
$$

The best-case scenario is that the alert prevents all future outcomes, i.e. that $\sum_{t=T_W}^T \Yt{f} = 0$. Under this assumption, the difference in expected outcomes is

$$
nE\left[
    \sum_{t=T_W}^{T} \Yt{\vec 0} | Z=1
    \right]P(Z=1)
$$

However, we have already shown above that $Z = \sum_{0}^T f(\Xt{f})$ is actually equal to  $\tilde W = \sum_{0}^T f(\Xt{0})$ when all alerts after the first are ignored and that the expected value of this quantity is estimated by $n_{\tilde W=1}/n$. Furthermore, the expectation in this expression is the same as $E[\tilde Y | \tilde W=1]$, which is estimated by $n_{\tilde Y=1, \tilde W=1}/n_{\tilde W=1}$. Thus we have that an unbiased estimate of the expected number of prevented outcomes using a perfectly effective intervention among $n$ patients when using $f$ to trigger alerts, and silencing alerts after the first alert or outcome, is

$$
n 
\left(
    \frac{n_{\tilde Y=1, \tilde W=1}}
        {n_{\tilde W=1}}
\right) 
\left(
    \frac{n_{\tilde W=1}}
        {n}
\right)
= n_{\tilde Y=1, \tilde W=1}$$

Since the intervention in practice will not be perfectly effective, this represents an upper bound on the number of outcomes that could be prevented with the alert system configured as specified. It is also possible to assume $\sum_{t=T_W}^T \Yt{f} \ne 0$ (i.e. an intervention that isn't perfect) and arrive at an estimate of the number of prevented outcomes. For instance, we could assume a particular risk ratio $\sum_{t=T_W}^T \Yt{f} = \rho \sum_{t=T_W}^T Y_{t,0}$, which implies that the alert prevents a fraction $\rho$ of all future outcomes that otherwise would have occurred. However, without external basis for making this assumption, the result would not be meaningfully interpretable as a bound. Despite that, this estimate could prove useful as a sensitivity analysis.

It is not sensible to use a look-ahead prediction window in the first alert evaluation. While it's easy to define a period of time after each alert, defining where that window should be for patients who did not trigger an alert is ambiguous and could not be done prospectively. 

In conclusion, the first alert evaluation can accurately estimate the number of alerts that would be produced by a system that runs continuously until an alert or outcome occurs, after which it is switched off. It can also estimate an upper bound on the number of outcomes that such a system could conceivably prevent, but, as with the fixed alert evaluation, that bound is of questionable utility for comparison between models if the intervention is time-sensitive.

\section{Example: Simulated Data}

To make these ideas concrete we will use simulated data in which we can recreate both the model building and evaluation process from non-interventional retrospective data as well as a (simulated) prospective randomized trial to perform a final evaluation. 

\subsection{Setup}

Our simulated ``patients'' are in fact point masses translating on a line. All patients experience a constant rightward force (e.g. propulsion), as well as a force that varies randomly at each time point (e.g. a buffeting wind).  Each patient therefore moves back and forth along a line according to these forces. The patient covariates are their current position, velocity, and acceleration. Patients start at rest at the origin and are considered to have an ``outcome'' when they have crossed more than one unit to the right, after which point they are frozen. Patients can therefore only experience a single outcome. We also assume the existence of an intervention that, when applied, imparts a strong leftward force at the time it is applied. The intervention can thus come too late if a patient is already moving rightward with too much momentum.

We used this simulation to generate a training dataset of 500 patients, each with 20 timepoints worth of data, and test dataset with the same number of patients and timepoints per patient. No interventions were applied in the generation of either of these datasets. We used the training data to fit a logistic regression model to predict whether the patient would suffer an outcome in the next 5 timepoints based on the current covariates. We then used risk cutoffs of 0.2, 0.4, 0.6, and 0.8 to generate four different alert models, and configured them so that no alerts would fire if the patient had already experienced either an alert or outcome.

\begin{table}[ht]
    \centering
\begin{tabular}{|l|r|r|r|}
\toprule
      Evaluation &  Threshold &  True Positives &  Positives \\
\midrule
 Aggregated Time &        0.2 &             289 &        908 \\
      Fixed Time &        0.2 &              31 &         42 \\
     First Alert &        0.2 &             702 &        908 \\
 Aggregated Time &        0.4 &             324 &        435 \\
      Fixed Time &        0.4 &              32 &         32 \\
     First Alert &        0.4 &             348 &        435 \\
 Aggregated Time &        0.6 &              96 &        136 \\
      Fixed Time &        0.6 &              11 &         11 \\
     First Alert &        0.6 &              98 &        136 \\
 Aggregated Time &        0.8 &              13 &         21 \\
      Fixed Time &        0.8 &               1 &          1 \\
     First Alert &        0.8 &              13 &         21 \\
\bottomrule
\end{tabular}
    \caption{Simulated model evaluation on retrospective data with a partially-effective intervention}
    \label{tab-eval}
\end{table}

\begin{table}[ht]
    \centering
\begin{tabular}{|r|r|r|}
\toprule
 Threshold &  Prevented Outcomes &  Alerts \\
\midrule
       0.2 &                 606 &     907 \\
       0.4 &                  74 &     495 \\
       0.6 &                   0 &     152 \\
       0.8 &                  27 &      24 \\
\bottomrule
\end{tabular}
    \caption{Simulated randomized trial results evaluating alert models with a partially-effective intervention}
    \label{tab-rct}
\end{table}

We then applied each of these four alert models to the test set to generate the alerts that would have been observed had these models been running. Of course, these models were not actually running when the data were generated, meaning that the ``observed'' alerts in the test set are virtual and did not actually have any effect on the generation of subsequent data. For each model, we then calculated the counts in the confusion matrices corresponding to the aggregated time, fixed time, and first alert evaluations. We used an outcome lookahead window of 5 timepoints for the aggregated time evaluation and set $t^* = 10$ for the fixed time evaluation. We report the numbers of true positives and positives for each of these in table \ref{tab-eval}.

Finally, we used each of these models (and a null model that never triggers alerts) in prospective simulations to generate the data that would have been observed if the model were running live and controlling the delivery of the intervention. This simulates a five-armed randomized trial with one model in each arm, plus the null control. We assigned 1000 patients to each arm of our simulated trial. We report the difference in the number of outcomes and the number of alerts between each model arm and the control arm in table \ref{tab-rct}.

\subsection{Results}

Firstly, we observe that the number of true positives in the first alert evaluation is indeed an estimated upper bound for the number of prevented outcomes. It is, however, an \textit{estimate}, and in some cases can be exceeded, as is the case in our simulation when the risk threshold is 0.8. This is in contrast to the number of true positives in the aggregated time evaluation, which sometimes is much larger than the number of prevented outcomes, and sometimes much smaller. When it is smaller (e.g. with the 0.2 threshold), it is because the aggregated time evaluation is using a lookahead window and thus cannot measure any potential effect on outcomes in the far future. On the other hand, there is also no guarantee that the intervention is effective in the short term, and thus it is also possible to overestimate the number of prevented outcomes. Not knowing which of these effects will win out is what makes the number of true positives in the aggregated time evaluation impossible to interpret, whereas in the first alert evaluation, we can be sure that the number of true positives is an overestimate of the number of prevented outcomes.

The number of positives in both the aggregate time and first alert evaluations are good estimates of the alert burdens prospectively encountered in our simulated trials. The two numbers of positives are the same because we structured our alert system to turn off after firing a single alert. If multiple alerts were possible, the number of positives in the aggregated time evaluation would increase, while the number of positives in the first alert evaluation would stay the same. The number of alerts in the prospective trial would also be expected to increase, but not necessarily by as much as would be implied by the aggregated time evaluation. It could easily be the case that a real alert decreases the subsequent possibility of alerts, which could make the number of positives an overestimate. The opposite could also be the case. The first alert evaluation avoids this issue by claiming that the number of positives estimates only the number of first alerts (i.e. the number of patients who ever experience any alert), which will at least be a consistent underestimate of the total number of alerts.

Finally, the numbers estimated in the fixed time evaluation seem to bear no relationship to the number of prevented outcomes or the number of observed alerts. This is to be expected- if our simulated randomized trial had been structured to apply a model at a pre-defined point in time (e.g. $t^*=10$) and only count outcomes after that point, the fixed time evaluation would have provided the appropriate upper bound on prevented outcomes. This underscores the fact that these methods are not competing to estimate the same thing; they estimate different quantities altogether.

\section{Example: Advanced Alert Monitor}

In addition to our simulated data, we also applied the risk model from \citet{kipnis2016} (the "Advanced Alert Monitor") to generate risk scores for a large test set designed as follow-up for that study. The data consist of N inpatient encounters over the course of NT years. Each patient in the dataset is assigned a risk score at each hour of their stay by an existing risk model trained to estimate their risk clinical deterioration (unplanned transfer to the ICU or in-hospital ward death in ``full-code'' patients) in the next 12 hours.

We used the risk scores generated by their model to build four alert systems based on thresholds of 0.01, 0.02, 0.03, and 0.04 and generated the virtual alerts that would have been observed had the model been running silently in the background. We then applied the three evaluation strategies to these data to generate counts of true positives and positives according to each approach. We used a lookahead window of 12 for the aggregated time evaluation and
$t^*=4$ (hours from admission) for the fixed time evaluation. 

\begin{table}[ht]
    \centering
    \begin{tabular}{|c|c|c|c|}
         \hline
         Evaluation & Threshold & True Positives & Positives \\
         \hline
         Aggregated Time & 0.01 & 87320 & 3637960\\
         Aggregated Time & 0.02 & 54127 & 1272158\\
         Aggregated Time & 0.03 & 38164 & 656134 \\
         Aggregated Time & 0.04 & 28879 & 406987\\
         \hline
         Fixed Time & 0.01 & 2583 & 93997\\
         Fixed Time & 0.02 & 1657 & 37009\\
         Fixed Time & 0.03 & 1165 & 19859\\
         Fixed Time & 0.04 & 904  & 12591 \\
         \hline
         First Alert & 0.01 & 14626 & 248356\\
         First Alert & 0.02 & 11728 & 131296 \\
         First Alert & 0.03 & 9798 & 85388\\
         First Alert & 0.04 & 8351 & 61753\\
         \hline
    \end{tabular}
    \caption{Evaluations of risk model from \citet{kipnis2016}}
\end{table}
%AASHISH ADDITIONS ON DATA: 
Since these are retrospective data, we cannot compare these results to the results we would have obtained by using these alert models in different arms of a randomized trial. However, these data are useful because they illustrate how these evaluation strategies can be applied to a large, real-world dataset. Furthermore, the results are clearly divergent between the different evaluation strategies in ways that mirror the divergences we observe in simulated data. 

% \section{Simulations}

% Some simulations to dig into and better demonstrate points that aren't communicated well by the example or the preceding math?

% \begin{itemize}
%     \item show how benefit doesn't correlate with number of TP when there is heterogeneity of effect in the single prediction case
%     \item count TPs and Ps using agg time, fixed time at t0, and first alert TO: first alert continuous, run once at t0, multiple alerts UNDER: a) alert prevents all subsequent outcomes (best case) b) some treatment effect
%     \item show how comparing bounds isn't necessarily useful (i.e. bounds may be similar but actual utility v. different?)
%     \item show how prediction window breaks down in the fixed time approach (i.e. outcomes actually prevented further out)
% \end{itemize}

\section{Discussion}

If only retrospective data is available and the intervention is not represented in the data, we recommend that practitioners employ the first alert evaluation method to evaluate repeated predictions models. Our argument is not that it ``works better'' than other methods for estimating the same thing, but that it estimates a quantity (or bounds on a quantity) that is actually of interest, whereas other approaches do not. Without assumptions, typical metrics of machine learning model performance calculated from retrospective data have no relation in theory to clinical utility. Using metrics based on confusion matrices to evaluate repeated predictions models is akin to fitting a square peg in a round hole.

The fundamental problem with the use of non-interventional retrospective data for model evaluation is that these data contain no information on how a putative intervention might affect outcomes. In single-prediction models, alert burdens are easy to estimate and assuming a risk ratio that is constant across the population makes the number of true positives in retrospective data a good predictor of the number of outcomes that would be prevented prospectively (although this assumption may not be warranted). Things are more complicated in repeated-prediction settings, but if the alert system is structured so that it would be turned off after the first alert or outcome, it is possible to estimate the alert burden and an upper bound on the number of outcomes that could be prevented. More targeted estimates of the number of prevented outcomes are possible, but only under strong assumptions about the effect of the intervention and how it varies in time. 

The conclusion is that a predictive model (or any model used to trigger an intervention) cannot be thoroughly evaluated using non-interventional retrospective data. Evaluations without interventional data either do not estimate meaningful quantities, require strong assumptions, or are limited to estimating best-case scenario bounds.

We recommend considering the intervention that the model is meant to target before evaluating the model. If the intervention is known, and represented in retrospective data, \textit{off-policy evaluation} methods can be used to estimate, e.g. the anticipated reduction in outcomes from using a given model to target the intervention \cite{athey2017state, liumodel, farajtabar2018more, kallus2018balanced}. Moreover, it is possible to train models to optimize these metrics in the first place using \textit{policy learning} methods instead of training models that predict risk of an outcome. In effect, risk modeling is the wrong framework altogether for building early-warning systems or model-triggered alerts when the intervention or a proxy thereof is available in retrospective data.

Risk models may still have utility as a composite of information about a patient, similar to a lab test or a traditional acuity score. But without very careful design or prospective evaluation, they should not be assumed to optimize the allocation of an intervention, no matter how well they perform in terms of standard predictive metrics. This is especially true in time-varying settings, where the assumptions necessary to link standard metrics to clinical utility are usually complex and unverifiable.

\section*{acknowledgements}
The authors thank John Greene, Patricia Kipnis, and Gabriel Escobar for providing data access and useful conversations. 

% \section*{conflict of interest}
% You may be asked to provide a conflict of interest statement during the submission process. Please check the journal's author guidelines for details on what to include in this section. Please ensure you liaise with all co-authors to confirm agreement with the final statement.

% \printendnotes

% Submissions are not required to reflect the precise reference formatting of the journal (use of italics, bold etc.), however it is important that all key elements of each reference are included.
\bibliography{references}

% \begin{biography}[example-image-1x1]{A.~One}
% Please check with the journal's author guidelines whether author biographies are required. They are usually only included for review-type articles, and typically require photos and brief biographies (up to 75 words) for each author.
% \bigskip
% \bigskip
% \end{biography}

% \graphicalabstract{example-image-1x1}{Please check the journal's author guildines for whether a graphical abstract, key points, new findings, or other items are required for display in the Table of Contents.}

\end{document}